\documentclass[journal]{IEEEtran}
\usepackage{hyperref}
\usepackage{times}
\usepackage{mathrsfs}
\usepackage{amssymb}
\usepackage{amsmath}
\usepackage{cases}
\usepackage{cite}
\usepackage{algorithm}
\usepackage{algorithmicx}
\usepackage{algpseudocode}
\usepackage{rotating}
\usepackage[inline]{enumitem}
\usepackage{multirow}
\usepackage{array}
\usepackage{subfigure}
\DeclareMathOperator*{\argmax}{argmax}
\DeclareMathOperator*{\argmin}{argmin}
\newtheorem{theorem}{Theorem}
\newtheorem{remark}{Remark}
\newtheorem{lemma}{Lemma}

\ifCLASSINFOpdf
  
\else
\fi
\begin{document}
%
\title{A Parameter-Free Learning Automaton Scheme}
%
%
%

\author{Hao~Ge,
			Shenghong~Li,~\IEEEmembership{Senior Member,~IEEE,}
			Jianhua~Li,
        and~Xudie~Ren
\thanks{This research work is funded by the National Science Foundation of China (61271316), Key Laboratory for Shanghai Integrated Information Security Management Technology Research, and Chinese National Engineering Laboratory for Information Content Analysis Technology.}
\thanks{H. Ge, S. Li, J. Li, and X. Ren are with the Department
of Electronic Engineering, Shanghai Jiao Tong University, Shanghai 200240, China (e-mail: sjtu\_gehao@sjtu.edu.cn; shli@sjtu.edu.cn; lijh888@sjtu.edu.cn; renxudie@sjtu.edu.cn}
\thanks{Manuscript received June 8, 2016; revised August 26, 2015.}}

\maketitle
\IEEEpeerreviewmaketitle
\begin{abstract}
For a learning automaton, a proper configuration of its learning parameters, which are crucial for the automaton's performance, is relatively difficult due to the necessity of a manual parameter tuning before real applications. To ensure a stable and reliable performance in stochastic environments, parameter tuning can be a time-consuming and interaction-costing procedure in the field of LA. Especially, it is a fatal limitation for LA-based applications where the interactions with environments are expensive.

In this paper, we propose a parameter-free learning automaton scheme to avoid parameter tuning by a Bayesian inference method. In contrast to existing schemes where the parameters should be carefully tuned according to the environment, the performance of this scheme is not sensitive to external environments because a set of parameters can be consistently applied to various environments, which dramatically reduce the difficulty of applying a learning automaton to an unknown stochastic environment. A rigorous proof of $\epsilon$-optimality for the proposed scheme is provided and numeric experiments are carried out on benchmark environments to verify its effectiveness. The results show that, without any parameter tuning cost, the proposed parameter-free learning automaton (PFLA) can achieve a competitive performance compared with other well-tuned schemes and outperform untuned schemes on consistency of performance.
\end{abstract}


%
\IEEEpeerreviewmaketitle

\section{Introduction}
%
%
%
%
\label{sec:introduction}
  \IEEEPARstart{L}{earning} Automata (LA) are simple self-adaptive decision units that were firstly investigated to mimic the learning behavior of natural organism\cite{narendra1974learning}.
  The pioneer work can be traced back to 1960s by the Soviet scholar Tsetlin\cite{tsetlin1961behavior, tsetlin1973automaton}.
  Since then, LA has been extensively explored and it is still under investigation as well in methodological aspects\cite{agache2002generalized, papadimitriou2004new, zhang2013incorporating, zhang2014last, ge2015novel, jiang2015new} as in concrete applications\cite{yang2007stochastic, oommen2010modeling,  horn2010solving, cuevas2012circle, yazidi2013learning, misra2014learning, kumar2015collaborative, vahidipour2015learning}.
  One intriguing property that popularize the learning automata based approaches in engineering is that LA can learn the stochastic characteristics of the external environment it interacts with, and maximize the long term reward it obtains through interacting with the environment.
  For a detailed overview of LA, one may refer to a new comprehensive survey in \cite{oommen2009cybernetics} and a classic book\cite{narendra2012learning}.

\begin{figure}[h]
  \includegraphics[width=3in]{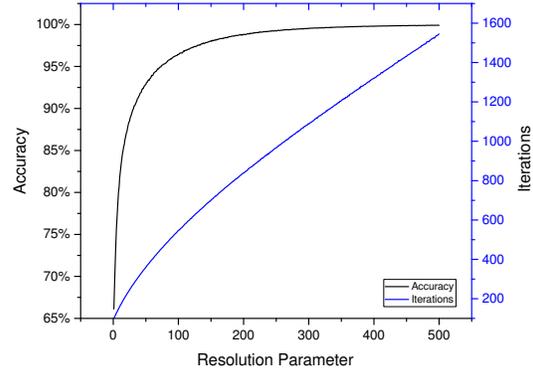}
\caption{The accuracy and iterations with different resolution parameters for DP$_{ri}$\cite{oommen1990discretized} in benchmark environment $E_1$, which is defined in \cite{papadimitriou2004new}\protect\footnotemark[2]. The results are averaged over 250000 replications.}
\label{fig:dilemma}
\end{figure}

  In the case of LA, \emph{accuracy} and \emph{convergence rate} becomes two major measurements to evaluate the effectiveness of a LA scheme.
  The former is defined as the probability of a correct convergence and the latter as the average iterations for a LA to get converged\footnotemark[1]\footnotetext[1]{For this reason, the terms \emph{convergence rate} and \emph{iteration} are used interchangeably.}. Most of the reported schemes in the field of LA has two or more tunable parameters, making themselves capable of adapting to a particular environment. The accuracy and convergence rate of an automaton are highly dependent on the selection of those parameters. Generally, ensuring a high accuracy is of uppermost priority. According to the \emph{$\epsilon$-optimality} property of LA, the probability of converging to the optimal action can be arbitrarily close to one, as long as the learning resolution is large enough. However, it will raise another problem. Taking the classic Pursuit scheme for example, as Fig.\ref{fig:dilemma} illustrates, the number of iterations required for convergence grows nearly linearly with the resolution parameter, while the accuracy grows logarithmically. This implies a larger learning resolution can lead to a higher accuracy, but at the cost of much more interactions with the environment. This dilemma necessitates parameter tuning to find a balance between convergence rate and accuracy.

  \footnotetext[2]{$E_1$ defined in \cite{papadimitriou2004new} corresponds to $E_5$ defined in section \ref{sec:results} of this paper.}
  
   In literatures, the performance of various LA schemes are evaluated by comparing their convergence rates on the premise of a certain accuracy. The learning parameters of various schemes are tuned through a standard procedure to ensure the accuracies are kept at the same level, so that the convergence rates can be fairly compared. For deterministic estimator based learning automata, the smallest value of the resolution parameter that yielded a hundred percent accuracy in a certain number of experiments is selected. The situation is more sophisticated when concerning the stochastic estimator based schemes\cite{papadimitriou2004new,ge2015novel,jiang2015new}, because extra configurable parameters should be set to control the perturbation added. Parameter tuning is intend to balance the trade-off between speed and accuracy. However, the interaction cost can be tremendous itself\footnotemark[3]\footnotetext[3]{The details will be elaborated in section \ref{sec:results}}, due to its trial and error nature. In practical applications, especially where interacting with environments could be expensive, e.g. drug trials, destructive tests and financial investments, the enormous cost for parameter tuning is undesired. Therefore, we believe, the issue of learning parameter configurations deserves more attention in the community, which give impetus to our work.

The scope of this research is confined to designing a learning scheme for LA in which the parameter tuning can be omitted, and that's why it is called \emph{parameter-free} in the title. It is noted that the term \emph{parameter-free} does not imply that no configurable parameters are involved in the proposed model, but indicates a set of parameters for the scheme can be universally applicable for all environments. This paper is an extension of our preliminary work\cite{ge2015parameter}. The proposed scheme in \cite{ge2015parameter} can only operate in two-action environments, whereas in this paper, our proposed scheme can operate in both two-action environments as well as multi-action environments. In addition, in this paper, optimistic initial values are utilized to improved the performance further. Moreover, a rigorous theoretical analysis of the proposed scheme and a comprehensive comparison among recently proposed LA schemes are provided in this paper which were not included in \cite{ge2015parameter}.

The contribution of this paper can be summarized as follows:
\begin{enumerate}
\item To the best of our knowledge, we present the first \emph{parameter-free} scheme in the field of LA, for learning in any stationary P-model stochastic environments. The meaning of the terminology \emph{parameter-free} is two-fold: (1) The learning parameters do not need to be manually configured. (2) Unlike other estimator based schemes, initializations of estimators are also unnecessary in our scheme. 
\item Most conventional LA schemes in literatures employ a stochastic exploration strategy, on the contrary, we design a deterministic gradient descent like method instead of probability matching as the exploration strategy to further accelerate the convergence rate of the automaton.
\item The statistics behavior of the proposed parameter-free learning automata (PFLA) is analyzed and a rigorous proof of the $\epsilon$-optimality property is provided as well.
\item A comprehensive comparison among recently proposed LA schemes is given to validate the theoretical analyses and demonstrate that PFLA is superior to other methods with respect to tuning cost.
\end{enumerate}

  This paper proceeds as follow.
  Section~\ref{sec:relate} describes our philosophy and some related works.
  Section~\ref{sec:algorithm} presents the primary results of the paper: a parameter-free learning automaton scheme.
  Section~\ref{sec:analysis} discusses the theoretic performance of the proposed scheme.
  Section~\ref{sec:results} provides a numerical simulation for verifying the proposed scheme.
  Finally, Section~\ref{sec:conclusion} concludes this paper.

\section{Related Works}
\label{sec:relate}
Consider a P-model environment which could be mathematically defined by a triple $\mathbb{<A,B,C>}$, where
\begin{itemize}
\item $\mathbb{A}=\{a_1,a_2,\dots,a_r\}$ represents a finite action set
\item $\mathbb{B}=\{0,1\}$ denotes a binary response set
\item $\mathbb{C}=\{c_1,c_2,\dots,c_r\}$ is a set of reward probabilities corresponding to $\mathbb{A}$, which means Pr\{$a_i$ gets rewarded\}=$c_i$. Each $c_i$ is assumed to lie in the open interval $(0,1)$.
\end{itemize}

Some other major notations that used throughout this paper are defined in table \ref{tab:1}.

\begin{table}
\caption{Notations used in this paper}
\label{tab:1}       
\begin{tabular}{cp{0.35\textwidth}}
\hline\noalign{\smallskip}
Symbol & Explanation  \\
\noalign{\smallskip}\hline\noalign{\smallskip}
$r$ & cardinality of the action set $\mathbb{A}$\\
$E$ & a vector of estimates\\
$N$ & the number of repetitions in Monte Carlo simulation\\
$\eta$ & the threshold to terminate the iteration\\
$a_i$ & the $i^{\text{th}}$ action in $\mathbb{A}$\\
$\alpha_i$ & a parameter of $a_i$'s beta distribution\\
$\beta_i$ & a parameter of $a_i$'s beta distribution\\
$S_i$ & the number of times that $a_i$ has been selected\\
$\mathcal{H}_i$ & the hypothesis that $a_i$ is the optimal action\\
$Beta(\alpha,\beta)$ & a beta distribution with parameter $\alpha$ and $\beta$ \\
$Norm(\mu,\sigma)$ & a normal distribution with mean $\mu$ and variance $\sigma^2$\\
$B(\alpha,\beta)$ & the beta function\\
$B(x;\alpha,\beta)$ & the incomplete beta function\\
\noalign{\smallskip}\hline
\end{tabular}
\end{table}

The aim of LA is to identify the optimal action $a_m$, which has the maximum reward probability, from $\mathbb{A}$ through interacting with the environment. A general philosophy is to collect feedbacks from environment and use these information to extract evidences that support an optimal assertion.

Then we are faced with two challenges:

\subsection{How to organize the information we gathered and  make full use of them?}
Lots of works have been done for the first challenge. Although the reward probabilities $\mathbb{C}$ is unknown to us, we can construct consistent estimators to guarantee that the estimates of the reward probabilities can converge to their true values as the quantity of samples increases.

As the feedback for one action can be modeled as a Bernoulli distributed random variable in P-model environment, there are two ways to construct such estimators currently.

\begin{enumerate}
\item One is from frequentist's perspective. The most intuitive approach is to utilize the likelihood function, which is a basic quantitative measure over a set of predictions with respect to observed data. In the context of parameter estimation, the likelihood function is naturally viewed as a function of the parameters $c_i$ to be estimated. The parameter that maximizes the likelihood of the observed data is referred to as \emph{maximum likelihood estimate} (MLE). MLE-based LA \cite{oommen1990discretized, agache2002generalized} are proved to be a great success, achieving a tremendous improvement on the rate of convergence comparing with traditional variable structure stochastic automata. However, as we revealed in \cite{ge2015novel}, MLE suffers from one principle weakness, i.e., MLE is unreliable when the quantity of samples is small. 

Several efforts has been devoted to improving MLE. The concept of \emph{stochastic estimator} was employed in \cite{papadimitriou2004new} so that the influence of lacking of samples can be reduced by introducing controlled randomnesses to MLE. In \cite{ge2015novel}, we proposed an interval estimator based learning automata DGCPA, in which the upper bound of a 99\% confidence interval of $c_i$ is used as estimates of reward probabilities. Both of these two LA schemes broke the records of convergence rate when proposed, which confirmed the defect of traditional MLE.

\item On the other hand, there are attempts from Bayesian perspective. Historically, one of the major reasons for avoiding Bayesian inference is that it can be computationally intensive under many circumstances. The rapid improvements in available computing power over the past few decades can, however, help overcome this obstacle, and Bayesian techniques are becoming more widespread not only in practical statistical applications but also in theoretical approaches to modeling human cognition. In Bayesian statistics, parameter estimation involves placing a probability distribution over model parameters. With regard to LA, the posterior distribution of $c_i$ with respect to observed data is a beta distribution.

In \cite{zhang2013incorporating}, DBPA was proposed where the posterior distribution of estimated $\hat{c_i}$ is represented by a beta distribution $Beta(\alpha,\beta)$, the parameter $\alpha$ and $\beta$ record the number of times that a specific action has been rewarded and penalized respectively. Then the $95^{\text{th}}$ percentile of the cumulative posterior distribution is utilized as estimation of $c_i$.
\end{enumerate}

One of main drawbacks of the way that information been used by existing LA schemes is that they summarize beliefs about $c_i$, such as the likelihood function or the posterior distribution, into a point estimate, which obviously may lead to information loss. In the proposed PFLA, we insist on taking advantage of the entire Bayesian posterior distribution of $c_i$ for further statistical inference.

\subsection{When is the time to make an assertion that claims one of the actions is optimal?}
For the second challenge, as the collected information accumulates, we become more and more confidence to make an assertion. But when is the exact timing? 

The quantity of samples before convergence of existing strategies is indirectly controlled by its learning parameters. Actually, the LA is not aware of whether it have collected enough information or not, as a consequence, its performance completely rely on the manual configuration of learning parameters inevitably.
As far as we're concerned, there is no report describing a parameter-free scheme for learning in multi-action environments, and this research area remains quite open.

However, there are efforts from other research area that shed some light on this target. In \cite{granmo2010solving}, a Bayesian learning automaton (BLA) was proposed for solving two-armed Bernoulli bandit (TABB) problem . The TABB problem is a classic optimization problem that explores the trade-off between exploitation and exploration in reinforcement learning. One distinct difference between learning automata and bandit playing algorithms is the metrics used for performance evaluation. Typically, \emph{accuracy} is used for evaluating LA algorithms while \emph{regret} is usually used in bandit playing algorithms. Despite being presented with different objective, BLA is somewhat related to our study and inspired our work. Therefore, the philosophy of BLA is briefly summarized as follows: The BLA maintains two beta distributions as estimates of the reward probabilities for the two arms (corresponding to actions in LA field). At each time instance, two values are randomly drawn from the two beta distributions respectively. The arm with the higher random value is selected, and the feedback is utilized to update the parameter of the beta distribution associated with the selected arm. One advantage of BLA is that it doesn't involve any explicit computation of Bayesian expression. In \cite{granmo2010solving}, it has been claimed that BLA performs better than UCB-tuned, the best performing algorithm reported in \cite{Auer2002}.

Inspired by \cite{granmo2010solving}, we constructe the PFLA by using Bayesian inference to enable convergence self-judgment in this paper. In contrast to \cite{granmo2010solving}, however, the probability of each arm being selected must be explicitly computed to judge the convergence of the algorithm. In addition, due to the poor performance of probability matching, we developed a deterministic exploration strategy. The technical details are provided in the next section.

\section{A Parameter-Free Learning Automaton}
\label{sec:algorithm}

In this section, we introduce each essential mechanism of our scheme in detail.

\subsection{Self-Judgment}
Consider a P-model environment with $r$ available actions, as we have no prior knowledge about these actions, each of them is possible to be the optimal one. We refer to these $r$ possibilities as $r$ hypotheses $\mathcal{H}_1,\mathcal{H}_2,\dots,\mathcal{H}_r$ so that each hypothesis $\mathcal{H}_i$ represents the event that action $a_i$ is the optimal action.

As we discussed in section \ref{sec:relate}, the Bayesian estimates of each action's reward probability just intuitively are beta distributed random variables, denoted as $E=\{e_1,e_2,\dots,e_r\}$, where $e_i \sim Beta(\alpha_i,\beta_i)$.

Because the propositions $\mathcal{H}_1,\mathcal{H}_2,\dots,\mathcal{H}_r$ are mutually exclusive and collectively exhaustive, apparently we have $\sum_i{Pr(\mathcal{H}_i)}=1$. Therefore, we can simply make an assertion that $\alpha_i$ is the optimal action once $Pr(\mathcal{H}_i)$ is greater than some predefined threshold $\eta$. For this reason the explicit computation of $Pr(\mathcal{H}_i)$ is necessary here to make that assertion.

\subsubsection{Two-Action Environments}
In the two-action case, $Pr(\mathcal{H}_1)$ can be formulated in the following equivalent forms: 
\begin{align}
Pr(\mathcal{H}_1)&=Pr(e_1>e_2)\\
&=\sum_{i=0}^{\alpha_1-1}\frac{B(\alpha_2+i,\beta_1+\beta_2)}{(\beta_1+i)B(1+i,\beta_1)B(\alpha_2,\beta_2)}\\
&=\sum_{i=0}^{\beta_2-1}\frac{B(\beta_1+i,\alpha_1+\alpha_2)}{(\alpha_2+i)B(1+i,\alpha_2)B(\alpha_1,\beta_1)}\\
&=1-Pr(\mathcal{H}_2)\\
&=1-\sum_{i=0}^{\alpha_2-1}\frac{B(\alpha_1+i,\beta_1+\beta_2)}{(\beta_2+i)B(1+i,\beta_2)B(\alpha_1,\beta_1)}\\
&=1-\sum_{i=0}^{\beta_1-1}\frac{B(\beta_2+i,\alpha_1+\alpha_2)}{(\alpha_1+i)B(1+i,\alpha_1)B(\alpha_2,\beta_2)}
\end{align}

The above formulas can be easily implemented by a programming language with well defined log-beta function, thus the exact calculation of $Pr(\mathcal{H}_1)$ can be completed within $\mathcal{O}(\min(\alpha_1,\alpha_2,\beta_1,\beta_2))$. However, in multi-action cases, the closed-form of $Pr(\mathcal{H}_i)$ is too complex and it's somewhat computationally intensive to calculate it directly. So in our scheme, a Monte Carlo simulation is adopted for evaluating $Pr(\mathcal{H}_i)$ in multi-action environment.

\subsubsection{Multi-Action Environments}
Closed-form calculation of $Pr(\mathcal{H}_i)$ is feasible for small action-set, but it becomes much more difficult as the number of actions increases.

Monte Carlo methods are a broad class of computational algorithms that rely on repeated random sampling to obtain numerical results \cite{wiki}.

In multi-action environments, in order to evaluate $Pr(\mathcal{H}_i)$, an intuitive approach is to generate random samples from the $r$ beta distributions and count how often the sample from $Beta(\alpha_i,\beta_i)$ is bigger than any other samples. By that way, the following Monte-Carlo simulation procedure is proposed.

Suppose the number of simulation replications is $N$. Since $e_i$ follows $Beta(\alpha_i,\beta_i)$, let $x_i^n$ be one of the $r$ random samples at the $n^{\text{th}}$ replication.

Then, $Pr(\mathcal{H}_i)$ can be simulated as 
\begin{equation}
\label{montecarlosim}
\widehat{Pr}(\mathcal{H}_i)= \frac{1}{N}\sum_{n=1}^N I(x_i^n)
\end{equation}
where $I(x_i^n)$ is an indicator function such that
\begin{subnumcases}
{I(x_i^n)=}
  1 & if $x_i^n >x_j^n,\forall j\ne i$\\
  0 & otherwise
\end{subnumcases}

It is simple to verify that $\sum_i Pr(\mathcal{H}_i)=1$.

\subsection{Exploration Strategy}
  In conventional estimator-based learning schemes, which are the majority family of LA, a stochastic exploration strategy is employed. A probability vector for choosing each action is maintained in the automaton and be properly updated under the guidance of the estimator and environment feedback after every interaction. However, such a probability vector does not exist in our scheme. Instead, a vector of probabilities indicating the chance of each action being the best one is maintained in our scheme. 
  The exploration strategy in \cite{granmo2010solving} is the so-called \emph{probability matching}, which occurs when an action is chosen with a frequency equivalent to the probability of that action being the best choice.  In \cite{ge2015parameter}, we constructed a learning automata by adding an absorbing barrier to BLA and apply it as a baseline for comparison. The numerical simulation shows the low performance of probability matching strategy on designing parameter-free LA. Therefore, a novel deterministic exploration strategy is proposed accordingly to overcome this pitfall.

  Because $\max\{Pr(\mathcal{H}_i)\}>\eta$ is the stop criterion of our scheme, in order to pursue a rapid convergence, one straightforward and obvious approach is maximizing the expected increment of $\max\{Pr(\mathcal{H}_i)\}$ over the action set. 

\subsubsection{Two-Action Environments}
In two-action environments, if $Pr(\mathcal{H}_1)$ is greater than $Pr(\mathcal{H}_2)$, then we suppose action $a_1$ is more likely to be the optimal one, and thus attempt to find out the action that will lead to the maximal expected increment of $Pr(\mathcal{H}_1)$, or vice versa.

  We denote $Pr(\mathcal{H}_1)$ as $g(\alpha_1,\beta_1,\alpha_2,\beta_2)$, and the following recurrence relations are derived \cite{cook2005exact}:
\begin{align}
g(\alpha_1+1,\beta_1,\alpha_2,\beta_2)&=g(\alpha_1,\beta_1,\alpha_2,\beta_2)+h(\alpha_1,\beta_1,\alpha_2,\beta_2)/\alpha_1\\
g(\alpha_1,\beta_1+1,\alpha_2,\beta_2)&=g(\alpha_1,\beta_1,\alpha_2,\beta_2)-h(\alpha_1,\beta_1,\alpha_2,\beta_2)/\beta_1\\
g(\alpha_1,\beta_1,\alpha_2+1,\beta_2)&=g(\alpha_1,\beta_1,\alpha_2,\beta_2)-h(\alpha_1,\beta_1,\alpha_2,\beta_2)/\alpha_2\\
g(\alpha_1,\beta_1,\alpha_2,\beta_2+1)&=g(\alpha_1,\beta_1,\alpha_2,\beta_2)+h(\alpha_1,\beta_1,\alpha_2,\beta_2)/\beta_2
\end{align}
where $h(\alpha_1,\beta_1,\alpha_2,\beta_2)=\frac{B(\alpha_1+\alpha_2,\beta_1+\beta_2)}{B(\alpha_1,\beta_1)B(\alpha_2,\beta_2)}$.

Hence, given that action $a_1$ is chosen, the conditional expected increment of $Pr(\mathcal{H}_1)$ is:
\begin{align}
&\mathbb{E}[\Delta{Pr(\mathcal{H}_1)}\mid a_1\text{ is chosen}]\\
&=c_1\times h(\alpha_1,\beta_1,\alpha_2,\beta_2)/\alpha_1 - (1-c_1)\times h(\alpha_1,\beta_1,\alpha_2,\beta_2)/\beta_1\\
&=h(\alpha_1,\beta_1,\alpha_2,\beta_2)(c_1/\alpha_1-(1-c_1)/\beta_1)
\end{align} 
because $c_1$ is unknown to us, we can approximate the above equation as
\begin{align}
&\mathbb{E}[\Delta{Pr(\mathcal{H}_1)}\mid a_1\text{ is chosen}]\\
&\approx h(\alpha_1,\beta_1,\alpha_2,\beta_2)(\frac{\alpha_1}{\alpha_1+\beta_1}/\alpha_1-\frac{\beta_1}{\alpha_1+\beta_1}/\beta_1)\\
\label{eq:action1}&=0
\end{align}
By the same way, we have
\begin{equation}
\label{eq:action2}\mathbb{E}[\Delta{Pr(\mathcal{H}_1)}\mid a_2 \text{ is chosen}]\approx 0
\end{equation}

  \eqref{eq:action1} and \eqref{eq:action2} indicate that no matter which action is picked, the expected difference of $\max\{Pr(\mathcal{H}_i)\}$ will approximately be zero, which makes it difficult for us to make decisions.

  Our solution is to select the action that give the expected maximum possible increment to $\max\{Pr(\mathcal{H}_i)\}$, as we did in \cite{ge2015parameter}. More specifically, if $Pr(\mathcal{H}_1)$ is greater than $Pr(\mathcal{H}_2)$, then we try to find out the action that could probably lead to the expected maximal increment of $Pr(\mathcal{H}_1)$, that is
\begin{equation}
\label{eq:object}
\argmax_i \mathbb{E}[\max\{\Delta Pr(\mathcal{H}_1)\}\mid a_i \text{ is chosen}]
\end{equation}
otherwise we try to maximize
\begin{equation}
\argmax_i \mathbb{E}[\max\{\Delta Pr(\mathcal{H}_2)\}\mid a_i \text{ is chosen}]
\end{equation}

The events that can lead to increments of $Pr(\mathcal{H}_1)$ are ``\emph{action $a_1$ is selected and rewarded}'' and ``\emph{action $a_2$ is selected and punished}''. Hence the optimization objective of \eqref{eq:object} can be simplified as:
\begin{subnumcases}
{\label{eq:maximumIncrement}}
\frac{c_1}{\alpha_1}h(\alpha_1,\beta_1,\alpha_2,\beta_2) & $a_1$ is chosen\\
\frac{1-c_2}{\beta_2}h(\alpha_1,\beta_1,\alpha_2,\beta_2) & $a_2$ is chosen

\end{subnumcases}
By employing the Maximum Likelihood Estimate of $c_1$ and $c_2$, \eqref{eq:maximumIncrement} can be written as
\begin{subnumcases}{}
\frac{h(\alpha_1,\beta_1,\alpha_2,\beta_2)}{(\alpha_1+\beta_1)} & $a_1$ is chosen\\
\frac{h(\alpha_1,\beta_1,\alpha_2,\beta_2)}{(\alpha_2+\beta_2)} & $a_2$ is chosen
\end{subnumcases}
The same conclusion holds also for situation $Pr(\mathcal{H}_1)<Pr(\mathcal{H}_2)$.

As a result, the strategy adopted in two-action environments is selecting the action which has been observed less between the two candidate actions at every time instance, as \eqref{sim_desc} reveals. 
\begin{subnumcases}
{\label{sim_desc}}
  \argmin_{i}(\alpha_i+\beta_i) & when $S_1\ne S_2$ \\
  \text{randomly chosen} & when $S_1= S_2$
\end{subnumcases}

\subsubsection{Multi-Action Environments}
  In multi-action environments, the automaton has to distinguish the best action from the action set. Intuitively, we can maximize expected increment of $Pr(\mathcal{H}_i)$ over the selection of actions, however, the closed form of $Pr(\mathcal{H}_i)$ is complicated, making the exact solution computationally intractable.

  However, from an alternative perspective, the automaton only need to determine which is the best from the top two possibly optimal actions. That is, for the two actions which are most possible to be the optimal action, denoted as action $a_{i1}$ and action $a_{i2}$, we only have to maximize the probability $Pr(e_{i1}>e_{i2})$ or $Pr(e_{i2}>e_{i1})$, exactly the same as it in two-action environments. So we come to the conclusion that, in the proposed scheme, our exploration strategy is similar to  \eqref{sim_desc}.

\subsection{Initialization of Beta Distributions}
In our scheme, each estimation $e_i$ is represented by a beta distribution $e_i\sim Beta(\alpha_i,\beta_i)$. The parameters $\alpha_i$ and $\beta_i$ record the number of times that action $a_i$ has been rewarded and punished, respectively. 

At the beginning, as we know nothing about the actions, a non-informative (uniform) prior distribution is advised to infer the posterior distribution. So $\alpha_i$ and $\beta_i$ should be set identically to 1,  exactly the same as in \cite{granmo2010solving, zhang2013incorporating}.

However, as clarified in \cite{rlbook}, initial action values can be used as a simple way of encouraging exploration. The technique of \emph{optimistic initial values} is applied, which has been reported as a quite effective simple trick on stationary problems.

Therefore, in our scheme, the prior distribution is $Beta(2,1)$ for inferring the posterior distribution, i.e., all beta random variables are initialized as $\alpha_i=2,\beta_i=1$. 

The estimates of all actions' reward probability are intentionally biased towards 1. The impact of the bias is permanent, though decreasing over iterations. When an action has been sampled just few times, the bias contributes a large proportion to the estimate, thus further exploration is encouraged. By the time an action has been observed many times, the impact of the biased initial value is negligible. 

Finally, the overall process of PFLA is summarized in algorithm \ref{PFLA}.

\begin{algorithm}[h!]
\caption{Parameter-Free Learning Automaton}
\label{PFLA}
\begin{algorithmic}[1]
\Require $\eta$: a convergence threshold; $N$: the number of replications of Monte Carlo simulation
\State \textbf{Initial} $\alpha_i=2,\beta_i=1$  for $i=1,2,3,\dots,r$;
\Repeat
\State Evaluate the probability $\widehat{Pr}(\mathcal{H}_i)$  according to \eqref{montecarlosim} for each action $i=1,2,3,\dots,r$;
\State Choose the two actions with top two $\widehat{Pr}(\mathcal{H}_i)$, denoted as $a_{i1}$ and $a_{i2}$. If there are two or more action with identical maximum $\widehat{Pr}(\mathcal{H}_i)$, then choose two from them randomly.
\State Select one from $a_{i1}$ and $a_{i2}$ according to 
\begin{subnumcases}
{a_i=}
  a_{i1}&if $S_{i1}< S_{i2}$\notag\\
  a_{i2}&if $S_{i1}> S_{i2}$\notag\\
  \text{randomly chosen} &if $S_{i1}= S_{i2}$\notag
\end{subnumcases}
and interacts with the environment.
\State Receive a feedback from the environment and update the parameters of beta distributions for action $a_i$:
\begin{subnumcases}
{}
  \alpha_i=\alpha_i+1 & if a reward is received\notag\\
  \beta_i=\beta_i+1 & if a penalty is received\notag
\end{subnumcases}

\Until $\max\{\widehat{Pr}(\mathcal{H}_i)\}>\eta$
\end{algorithmic}
\end{algorithm}


\section{Performance Analysis}
\label{sec:analysis}

In this section, the statistical performance of the proposed scheme is analyzed, an approximate lower bound of the accuracy is derived and the $\epsilon$-optimality of the proposed scheme is further proved.

\subsection{An Approximate Lower Bound of the Accuracy}
As declared in \cite{mcbook}, from the central limit theorem (CLT), we know that the error of Monte Carlo simulation has approximately a normal distribution with zero mean and variance $\sigma^2/N$. Hence, if we denote the error between $Pr(\mathcal{H}_i)$ and its Monte-Carlo estimate as $\epsilon_i$, then we get
\begin{align}
Pr(\mathcal{H}_i)&=\widehat{Pr}(\mathcal{H}_i)+\epsilon_i\\
&\ge \eta+\epsilon_i\\
&\sim \eta+Norm(0,\frac{\sigma^2_i}{N})\label{eq:accuracy}\\
&\ge \eta-\mid \epsilon_i \mid
\end{align}
where $\widehat{Pr}(\mathcal{H}_i)$ is the Monte-Carlo estimate of $Pr(\mathcal{H}_i)$ and $\sigma^2_i$ is the variance of $I(x_i)$. 

We may note that the right hand side of \eqref{eq:accuracy} is irrelevant to the characteristics of the environment. In other words, the performance of the proposed scheme only depends on the selection of $\eta$ and $N$. That is the theoretical foundation of the parameter-free property.

\begin{figure}[h]
  \includegraphics[width=3in]{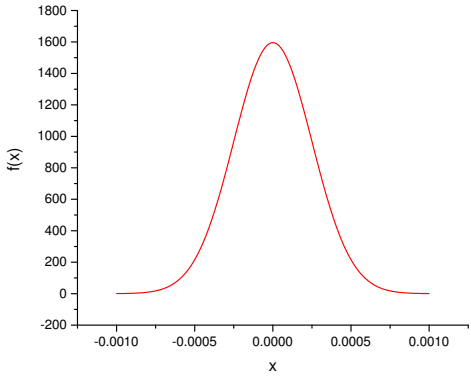}
\caption{The probability density function of $Norm(0,\frac{1}{4000})$}
\label{fig:normpdf}
\end{figure}

As the outcome of $I(x_i)$ is binary, in worst case, the maximum of $\sigma^2_i$ is 0.25. When N equals 1000, the probability density function of $\epsilon_i$ is shown in Fig.\ref{fig:normpdf}, which quantitatively depicts the error. Obviously, the error is so small that could be ignored.

Therefore, the approximate lower bound of $Pr(\mathcal{H}_i)$ is $\eta$. According to the Bayesian theory, the accuracy of our scheme is approximately larger than $\eta$.

Next, we shall describe the behavior of the proposed scheme more precisely. Like the pioneers have done in previous literatures, the $\epsilon$-optimality of the proposed scheme will be derived.

\subsection{Proof of $\varepsilon$-optimality}

Recall that $e_i$ is defined as the estimated reward probability of action $a_i$ and follows $Beta(\alpha_i,\beta_i)$, which is the posterior distribution of the estimated reward probability. The probability density function of $Beta(\alpha_i,\beta_i)$ is $f(x_i;\alpha_i,\beta_i)=C_ix_i^{\alpha_i-1}(1-x_i)^{\beta_i-1}$, where $C_i=\frac{1}{B(\alpha_i,\beta_i)}$ serves as a normalizing factor such that $\int_0^1 f(x_i;\alpha_i,\beta_i)=1$. Let $Z_i=\alpha_i-2$ and $W_i=\beta_i-1$ denote the numbers of times that action $a_i$ has been rewarded and penalized respectively, and $S_i=Z_i+W_i=\alpha_i+\beta_i-3$ be the the number of times that action $a_i$ has been selected.

Based on these preliminaries, the following Lemmas and Theorems are proposed:
\begin{lemma}
\label{ConvergeInProbability}
The beta distribution $Beta(\alpha_i,\beta_i)$ becomes 1-point Degenerate distribution with a Dirac delta function spike at $c_i$, provided that the number of selecting action $a_i$ approaches infinity, i.e. $\forall \varepsilon>0$,
\begin{align}
\lim_{S_i \to \infty} \int_{\mid x_i-c_i \mid\le \varepsilon \bigcap [0,1]}f(x_i;\alpha_i,\beta_i)dx_i=1\\
\lim_{S_i \to \infty} \int_{\mid x_i-c_i \mid> \varepsilon \bigcap [0,1]}f(x_i;\alpha_i,\beta_i)dx_i=0
\end{align}
\end{lemma}
\begin{IEEEproof}
According to the law of large numbers, we have $\frac{Z_i}{S_i}\to c_i$, as $S_i\to \infty$.

Hence
\begin{equation}
\left.
\begin{split}
&\lim_{S_i\to \infty} \frac{\alpha_i-1}{S_i}=\frac{Z_i+1}{S_i}=c_i\\
&\lim_{S_i\to \infty} \frac{\beta_i-1}{S_i}=\frac{S_i-Z_i}{S_i}=(1-c_i)
\end{split}
\right\}
\Rightarrow
\left\{
\begin{split}
\alpha_i-1&=c_iS_i\\
\beta_i-1&=(1-c_i)S_i
\end{split}
\right.
\end{equation}

The probability density function takes the form:
\begin{align}
\lim_{S_i\to \infty} f(x_i;\alpha_i,\beta_i) &=C_i x_i^{\alpha_i-1}(1-x_i)^{\beta_i-1}\\
&=C_i[x_i^{c_i}(1-x_i)^{1-c_i}]^{S_i}\\
&=C_ig^{S_i}(x_i)
\end{align}
where $g(x_i)=x_i^{c_i}(1-x_i)^{1-c_i}$.

Note that $g(x_i)$ is a nonnegative integrable function, we have
\begin{equation}
\lim_{S_i\to \infty} \left(\int_0^1 g^{S_i}(x_i)\right) ^{\frac{1}{S_i}} dx_i=||g||_{\infty}.
\end{equation}

Therefore,
\begin{equation}
\lim_{S_i\to \infty} C_i^{\frac{1}{S_i}}=\frac{1}{\left(\int_0^1 g^{S_i}(x_i)dx_i\right)^{\frac{1}{S_i}}}=\frac{1}{||g||_\infty}
\end{equation}

This reveals, as $S_i\to \infty$
\begin{align}
&\left(\int_{\mid x_i-c_i \mid> \varepsilon \bigcap [0,1]} f(x_i;\alpha_i,\beta_i)dx_i\right)^{\frac{1}{S_i}}\notag\\
=&C_i^{\frac{1}{S_i}}\left(\int_{\mid x_i-c_i \mid> \varepsilon \bigcap [0,1]} g^{S_i}(x_i)dx_i\right)^{\frac{1}{S_i}}\to \frac{\|g\|_{\infty,\varepsilon}}{\|g\|_\infty}
\label{norm}
\end{align}
where $\|g\|_{\infty,\varepsilon}$ is the $L^\infty$ norm of $g$ when restricted to ${|x_i-c_i|>\varepsilon}$.

By taking both sides of \eqref{norm} to the $S_i$ power, we obtain
\begin{equation}
\int_{\mid x_i-c_i \mid> \varepsilon \bigcap [0,1]} f(x_i;\alpha_i,\beta_i)dx_i \to \left(\frac{\|g\|_{\infty,\varepsilon}}{\|g\|_\infty}\right)^{S_i}
\end{equation}

Obviously $\frac{\|g\|_{\infty,\varepsilon}}{\|g\|_\infty}<1$, for the fact that $g$ is continuous and has a unique maximum at $c_i$, thus
\begin{equation}
\int_{\mid x_i-c_i \mid> \varepsilon \bigcap [0,1]} f(x_i;\alpha_i,\beta_i)dx_i \to 0
\end{equation}
as $S_i\to \infty$.

Note that $\int_0^1 f(x_i;\alpha_i,\beta_i)dx_i=1$ and the proof is finished.

\end{IEEEproof}

\begin{lemma}
\label{compare}
For two or more random variables $e_i \sim Beta(\alpha_i,\beta_i)$, assume $m$ is the index of action that has the maximum reward probability such that $c_m=\max(c_i)$, then
\begin{equation}
\lim_{S_i\to \infty} Pr\{e_m>\max_{i\ne m}(e_i)\}=1
\end{equation}
\end{lemma}
\begin{IEEEproof}
\begin{align}
\label{compareintegral}
&Pr\{e_m>\max_{i\ne m}(e_i)\} \notag\\
=& \int_0^1 f(x_m;\alpha_m,\beta_m) \prod_{i\ne m}[\int_0^{x_m} f(x_i;\alpha_i,\beta_i)dx_i] dx_m
\end{align}

From Lemma \ref{ConvergeInProbability}, we know that $f(x_i;\alpha_i,\beta_i)\to\delta(x_i-c_i)$ as $S_i\to \infty$.

By using the sampling property of Dirac delta function, \eqref{compareintegral} can be simplified as
\begin{align}
\lim_{S_i\to \infty} \Pr\{e_m>\max_{i\ne m}(e_i)\} &= \lim_{S_i\to \infty} \prod_{i\ne m}\int_0^{c_m} f(x_i;\alpha_i,\beta_i)dx_i\\
&=\lim_{S_i\to \infty} \prod_{i\ne m}\int_0^{c_m} \delta(x_i-c_i)dx_i
\end{align}

Note that $\forall i\ne m$, as $c_i\in[0,c_m]$, $\int_0^{c_m} \delta(x_i-c_i)dx_i=1$. And finally
\begin{equation}
\lim_{S_i\to \infty} Pr\{e_m>\max_{i\ne m}(e_i)\}=1
\end{equation}
This completes the proof.
\end{IEEEproof}

\begin{remark}
\label{correctconverge}
It is noted that, Lemma \ref{compare} implies $\lim_{S_i\to \infty} Pr\{\mathcal{H}_m\}=1$
\end{remark}

\begin{lemma}
\label{infinitysample}
Suppose one component of the vector $\{Pr(\mathcal{H}_1),Pr(\mathcal{H}_2),\dots,Pr(\mathcal{H}_r)\}$, say $Pr(\mathcal{H}_i)$ approaches 1 only if the number of each action been selected $S_i\to \infty$, for all $i\in\{1,2,\dots,r\}$.
\end{lemma}
\begin{IEEEproof}
As $Pr(\mathcal{H}_i)\to 1$,
for any $\delta>0$, we have $Pr(\mathcal{H}_i)\ge 1-\delta$, hence
\begin{align}
Pr(\mathcal{H}_i) =& \int_0^1 f(x_i;\alpha_i,\beta_i) \prod_{j\ne i}[\int_0^{x_i} f(x_j;\alpha_j,\beta_j)dx_j] dx_i\\
=& \int_0^1 f(x_i;\alpha_i,\beta_i) \int_0^{x_i} f(x_j;\alpha_j,\beta_j)dx_j \notag\\
&\prod_{k\ne i,k\ne j}[\int_0^{x_i} f(x_k;\alpha_k,\beta_k)dx_k] dx_i\\
\le& \int_0^1 f(x_i;\alpha_i,\beta_i) \int_0^{x_i} f(x_j;\alpha_j,\beta_j)dx_j \notag\\
&\prod_{k\ne i,k\ne j}[\int_0^1 f(x_k;\alpha_k,\beta_k)dx_k] dx_i\\
=& \int_0^1 f(x_i;\alpha_i,\beta_i) \int_0^{x_i} f(x_j;\alpha_j,\beta_j)dx_j dx_i\\
=&Pr\{e_i>e_j\}
\end{align}
As a result, for all $j\ne i$,
\begin{align}
&Pr\{e_i>e_j\}\ge Pr(\mathcal{H}_i)\to1\\
\Rightarrow &Pr\{e_j>e_i\}\to 0\\
\Rightarrow &\int_0^1 f(x_j;\alpha_j,\beta_j) \int_0^{x_j} f(x_i;\alpha_i,\beta_i)dx_idx_j \to 0
\label{zero-integral1}
\end{align}
By denoting $F(x)=f(x;\alpha_j,\beta_j)B(x;\alpha_i,\beta_i)=f(x;\alpha_j,\beta_j) \int_0^{x} f(x_i;\alpha_i,\beta_i)dx_i$, we have
\begin{equation}
\label{zero-integral2}
\int_0^1 F(x) dx \to 0
\end{equation}

Suppose at least one of $S_i$ and $S_j$ is not infinity, thus three possible cases should be discussed.
\begin{enumerate}
\item \emph{Case $S_i<\infty$ and $S_j<\infty$.}

In this case, $f(x_j;\alpha_j,\beta_j)$ is a continuous function and strictly positive on $(0,1)$. As $\frac{dB(x;\alpha_i,\beta_i)}{dx}=f(x_i;\alpha_i,\beta_i)$ is continuous, $B(x;\alpha_i,\beta_i)$ is continuously differentiable which implies it is a continuous function. In addition, $B(x;\alpha_i,\beta_i)$ is strictly positive on $(0,1]$. Clearly, the product of two strictly positive continuous functions $F(x)$ is continuous and $F(x)>0$ on the interval $(0,1)$, hence
\begin{equation}
\int_0^1 F(x)dx > 0
\end{equation}
which contradicts with \eqref{zero-integral2}.

\item Case $S_i<\infty$ and $S_j=\infty$.

Similarly, we can prove that $B(x;\alpha_i,\beta_i)$ is strictly positive and continuous on $(0,1]$, and $f(x;\alpha_j,\beta_j)\to \delta(x-c_j)$.

Hence, \eqref{zero-integral1} can be written as:
\begin{equation}
B(c_j;\alpha_i,\beta_i) \to 0
\end{equation}
that contradicts with the fact that $B(x;\alpha_i,\beta_i)$ is strictly positive on $(0,1]$.

\item Case $S_i=\infty$ and $S_j<\infty$.

Similarly, we can prove that $f(x_j;\alpha_j,\beta_j)$ is strictly positive and continuous on $(0,1)$, and $f(x_i;\alpha_i,\beta_i)\to \delta(x-c_i)$. 

Hence, \eqref{zero-integral1} can be written as:
\begin{equation}
\int_{c_i}^1 f(x;\alpha_j,\beta_j) dx \to 0
\end{equation}
which implies $f(x_j;\alpha_j,\beta_j)=0$ on $(c_i,1)$, that contradicts with the fact that $f(x_j;\alpha_j,\beta_j)$ is strictly positive on $(0,1)$.
\end{enumerate}

By summarizing the above three cases, we conclude that the supposition is false and both $S_i$ and $S_j$ must be infinity.

As $i,j$ enumerate all the action indexes, the proof is completed.
\end{IEEEproof}

\begin{remark}
\label{rem:2}
From Lemma \ref{infinitysample} and Remark \ref{correctconverge}, one can immediately see that given a threshold $\eta\to1$, PFLA will converge to the optimal action w.p.1 whenever it gets converged.
\end{remark}

\begin{lemma}
\label{montecarlo}
The Monte Carlo estimation of $Pr(\mathcal{H}_i)$ will converge almost surely as the number of Monte Carlo replications $N$ tends to infinity. i.e.:
\begin{equation}
Pr\{ \lim_{N\to\infty} \widehat{Pr}(\mathcal{H}_i) = Pr(\mathcal{H}_i)\}=1
\end{equation}
\end{lemma}
\begin{IEEEproof}
This lemma can be easily derived according to the strong law of large numbers.
\end{IEEEproof}

Let us now state and prove the main result for algorithm PFLA.

\begin{theorem}
PFLA is $\epsilon$-optimal in every stationary random environment. That is, given any $\varepsilon>0$, there exist a $N_0<\infty$, a $t_0<\infty$ and a $\eta_0<1$ such that for all $t\ge t_0$, $N\ge N_0$ and $\eta>\eta_0$:
\begin{equation}
\widehat{Pr}(\mathcal{H}_m)>1-\varepsilon
\end{equation}
\end{theorem}

\begin{IEEEproof}
The theorem is equivalent to show that,
\begin{equation}
\label{target}
Pr\{ \lim_{\substack{N\to\infty\\ t\to\infty\\ \eta\to1}} \widehat{Pr}(\mathcal{H}_m)=1\}=1
\end{equation}

From Lemma \ref{montecarlo}, we know that \eqref{target} is equivalent to
\begin{equation}
Pr\{ \lim_{\substack{t\to\infty\\ \eta\to1}} Pr(\mathcal{H}_m)=1\}=1
\end{equation}

And according to Remark \ref{rem:2}, we only need to prove that the scheme can definitely get converged, i.e., at least one of the components $\{Pr(\mathcal{H}_1),Pr(\mathcal{H}_2),\dots,Pr(\mathcal{H}_r)\}$ approaches 1, as $t\to\infty$ and $\eta\to1$.

Suppose the scheme have not converged yet at time $t_1$, because exactly one action will be explored at each time instant, we have $\sum_i S_i = t_1$.

As $t_1\to\infty$, a finite series has an infinite sum, which indicates that at least one of the terms $S_i$ has infinite value. 

Then denote the set of actions, whose corresponding observation times $S_i(t_1)\to\infty$, as $\mathbb{A}_1$, and denote the absolute complement set of $\mathbb{A}_1$ as $\mathbb{A}_2$.
\begin{enumerate}
\item If $\mathbb{A}_2 = \emptyset$, then for any action $a_i$, we have $S_i \to \infty$. 

By considering Remark \ref{correctconverge}, we have
\begin{equation}
Pr(\mathcal{H}_m) \to 1
\end{equation}

\item We will show that if $\mathbb{A}_2 \ne \emptyset$, then it is impossible that both the top two possibly optimal actions belong to set $\mathbb{A}_1$.

Denote the action in $\mathbb{A}_1$ with the highest reward probability as $a_{m1}$, then according to Lemma \ref{compare}, $\forall a_i\in \mathbb{A}_1$ and $i\ne m1$, 
\begin{equation}
Pr(\mathcal{H}_i) \to 0.
\end{equation} 

While for actions $a_j\in \mathbb{A}_2$,
\begin{align}
Pr(\mathcal{H}_j)=&\int_0^1 f(x_j;\alpha_j,\beta_j) \prod_{k\ne j}[\int_0^{x_j} f(x_k;\alpha_k,\beta_k)dx_k] dx_j\\
=&\int_0^1 f(x_j;\alpha_j,\beta_j) \prod_{a_{k1} \in \mathbb{A}_1}[\int_0^{x_j} f(x_{k1};\alpha_{k1},\beta_{k1})dx_{k1}] \notag\\
&\prod_{k2\ne i,a_{k2} \in \mathbb{A}_2}[\int_0^{x_j} f(x_{k2};\alpha_{k2},\beta_{k2})dx_{k2}] dx_j\\
=&\int_0^1 f(x_j;\alpha_j,\beta_j) \prod_{a_{k1} \in \mathbb{A}_1}I(x_j\ge c_{k1}) \notag\\
&\prod_{k2\ne i,a_{k2} \in \mathbb{A}_2}[\int_0^{x_j} f(x_{k2};\alpha_{k2},\beta_{k2})dx_{k2}] dx_j\\
=&\int_{c_{m1}}^1 f(x_j;\alpha_j,\beta_j) \notag\\
&\prod_{k2\ne i,a_{k2} \in \mathbb{A}_2}[\int_0^{x_j} f(x_{k2};\alpha_{k2},\beta_{k2})dx_{k2}] dx_j
\label{eqA2}
\end{align} 
As $c_{m1}<1$, and the integrand is strictly positive and continuous.
Obviously, \eqref{eqA2} is larger than zero trivially.

For actions in $\mathbb{A}_1$ other than $a_{m1}$, $Pr(\mathcal{H}_i)\to 0$, while for actions in $\mathbb{A}_2$, all $Pr(\mathcal{H}_i)$ equal some constants that are larger than zero. Hence, at least one action of the top two most probably optimal actions is from $\mathbb{A}_2$ and this action will be chosen to draw a feedback.

As time $t \to \infty$, once $\mathbb{A}_2 \ne \emptyset$, one action in $\mathbb{A}_2$ will be explored. As a consequence, we can always find a $t_0>t_1$ such that all actions in $\mathbb{A}_2$ will be explored infinite times and yield an empty $\mathbb{A}_2$.
\end{enumerate}

Combining the above two cases, we may infer that all actions will be explored with infinite number of times and $Pr(\mathcal{H}_m) \to 1$.

This completes the proof.
\end{IEEEproof}


\section{Simulation Results}
\label{sec:results}
During the last decade, SE$_{ri}$ has been considered as the state-of-art algorithm for a long time, however, some recently proposed algorithms \cite{ge2015novel,jiang2015new} claim a faster convergence than SE$_{ri}$. To make a comprehensive comparison among currently available techniques, as well as to verify the effectiveness of the proposed parameter-free scheme, in this section, PFLA is compared with several classic parameter-based learning automata schemes, including DP$_{ri}$\cite{oommen1990discretized}, DGPA\cite{agache2002generalized}, DBPA\cite{zhang2013incorporating}, DGCPA$^*$\cite{ge2015novel}, SE$_{ri}$\cite{papadimitriou2004new}, GBSE\cite{jiang2015new} and LELA$_{R}$\cite{zhang2014last}.

All the schemes are evaluated in four two-action benchmark environments\cite{ge2015parameter} and five ten-action benchmark environments\cite{papadimitriou2004new}. The actions' reward probabilities for each environment are as follows:
\begin{itemize}
\item[$E_1$]:$\{0.90,0.60\}$
\item[$E_2$]:$\{0.80,0.50\}$
\item[$E_3$]:$\{0.80,0.60\}$
\item[$E_4$]:$\{0.20,0.50\}$
\item[$E_5$]:$\{0.65,0.50,0.45,0.40,0.35,0.30,0.25,0.20,0.15,0.10\}$
\item[$E_6$]:$\{0.60,0.50,0.45,0.40,0.35,0.30,0.25,0.20,0.15,0.10\}$
\item[$E_7$]:$\{0.55,0.50,0.45,0.40,0.35,0.30,0.25,0.20,0.15,0.10\}$
\item[$E_8$]:$\{0.70,0.50,0.30,0.20,0.40,0.50,0.40,0.30,0.50,0.20\}$
\item[$E_9$]:$\{0.10,0.45,0.84,0.76,0.20,0.40,0.60,0.70,0.50,0.30\}$
\end{itemize}

The comparison is organized in two ways: 
\begin{enumerate*}
\item Comparison between PFLA and parameter-based schemes with their learning parameters being carefully tuned.
\item Comparison between PFLA and parameter-based schemes without parameter tuning, using either pre-defined or randomly selected learning parameters.
\end{enumerate*}

\subsection{Comparison with Well-tuned Schemes}

 Firstly, the parameter-based schemes are simulated with carefully tuned best parameters. The procedure for obtaining best parameters is elaborated in the Appendix. The proposed PFLA, by contrast, take identical parameter values of $\eta=0.99$ and $N=1000$ in all nine environments.

\begin{table}
\caption{Accuracy(\emph{number of correct convergences/number of experiments}) of the compared algorithms in environments $E_1$ to $E_9$, when using the `best' learning parameters(250,000 experiments were performed for each scheme in each environment)}
\label{tab:accuracy}       
\begin{tabular}{p{0.01\textwidth}*{8}{p{0.035\textwidth}}}
\hline\noalign{\smallskip}
Env.& DP$_{ri}$ & DGPA & DBPA & DGCPA$^*$ & SE$_{ri}$ & GBSE & LELA$_{R}$ & PFLA  \\
\noalign{\smallskip}\hline\noalign{\smallskip}
$E_1$ & 0.997 & 0.998 & 0.997 & 0.997 & 0.998 & 0.998 & 0.998 & 0.999 \\
$E_2$ & 0.997 & 0.997 & 0.998 & 0.998 & 0.997 & 0.998 & 0.998 & 0.999\\
$E_3$ & 0.996 & 0.996 & 0.996 & 0.997 & 0.997 & 0.997 & 0.997 & 0.998\\
$E_4$ & 0.998 & 0.997 & 0.998 & 0.997 & 0.998 & 0.998 & 0.998 & 0.999\\
$E_5$ & 0.995 & 0.997 & 0.996 & 0.997 & 0.997 & 0.997 & 0.997 & 0.997 \\
$E_6$ & 0.994 & 0.996 & 0.994 & 0.996 & 0.996 & 0.996 & 0.996 & 0.999 \\
$E_7$ & 0.993 & 0.995 & 0.993 & 0.995 & 0.995 & 0.995 & 0.995 & 0.996\\
$E_8$ & 0.996 & 0.997 & 0.996 & 0.998 & 0.998 & 0.998 & 0.997 & 0.999\\
$E_9$ & 0.994 & 0.997 & 0.994 & 0.997 & 0.997 & 0.997 & 0.997 & 0.997\\
\noalign{\smallskip}\hline
\end{tabular}
\end{table}

The results of simulations are summarized in Table \ref{tab:accuracy} and Table \ref{tab:iterations}. The \emph{accuracy} is defined as \emph{the ratio between the number of correct convergence and the number of experiments}, whilst the \emph{iteration} as \emph{the averaged number of required  interactions between automaton and environment for a correct convergence}. It is noted that the initialization cost of estimators is also included. The number of initializations for each action is 10.

In Table \ref{tab:accuracy}, PFLA converge with a relative high accuracy consistently, coinciding with our analytical results in Section \ref{sec:analysis}, and verifing the effectiveness of our proposed parameter-free scheme. And since the accuracies of all schemes are close, their convergence rates can be ``fairly'' compared\footnotemark[4]\footnotetext[4]{Technically speaking, the comparison in not completely fair, that's the reason the word ``fairly'' are quoted. Explanation will be given in later subsections.}.

\begin{figure}[h]
\subfigure[Two-action environments]{
\includegraphics[width=3in]{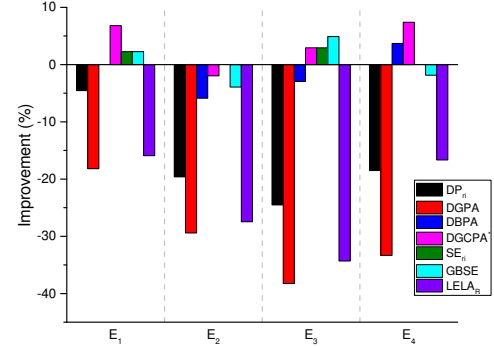}}
\subfigure[Ten-action environments]{
\includegraphics[width=3in]{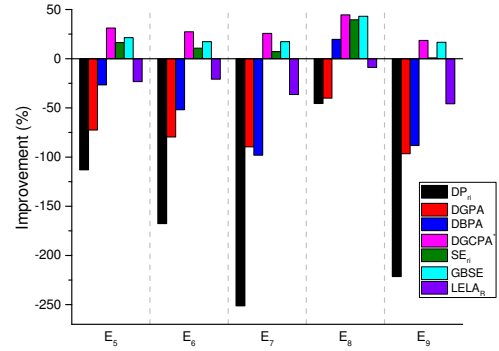}}
\caption{Convergence rate improvements of the compared algorithms relative to PFLA in benchmark environments, calculated by using $\frac{\text{Iterations}_{\text{\{PFLA\}}}-\text{Iterations}_{\text{\{ComparedAlgorithm\}}}}{\text{Iterations}_{\text{\{PFLA\}}}}$}
\label{fig:improvement}
\end{figure}

In the aspect of convergence rate, obviously in Table \ref{tab:iterations}, PFLA is outperformed by the top performers, namely SE$_{ri}$, GBSE and DGCPA$^*$. Figure \ref{fig:improvement} depicts the improvements of the competitors with respect to PFLA. Take $E_7$ as an example, the convergence rate of PFLA is improved by DGCPA$^*$, SE$_{ri}$ and GBSE with 25.76\%, 7.20\% and 17.35\%, respectively. While other schemes, DP$_{ri}$, DGPA and LELA$_R$ are outperformed by PFLA significantly. Generally speaking, FPLA is faster than deterministic estimator based schemes and slower than stochastic estimator based algorithms.

\begin{sidewaystable}
\caption{Comparison of the average number of iterations required for convergence of the compared algorithms in environments $E_1$ to $E_9$. For all schemes, the `best' learning parameters for each environment are used(250,000 experiments were performed for each scheme in each environment)}
\label{tab:iterations}       
\begin{tabular}{{p{0.01\textwidth}<{\centering}}*{6}{p{0.04\textwidth}<{\centering}}*{3}{p{0.07\textwidth}<{\centering}p{0.04\textwidth}<{\centering}}*{3}{p{0.04\textwidth}<{\centering}}}
\hline\noalign{\smallskip}
\multirow{2}{*}{\centering Env.} & \multicolumn{2}{c}{DP$_{ri}$} & \multicolumn{2}{c}{DGPA} & \multicolumn{2}{c}{DBPA} & \multicolumn{2}{c}{DGCPA*} & \multicolumn{2}{c}{SE$_{ri}$} & \multicolumn{2}{c}{GBSE}& \multicolumn{2}{c}{LELA$_{R}$}& {PFLA}  \\
& Para. & Iteration & Para. & Iteration & Para. & Iteration & Para. & Iteration & Para. & Iteration & Para. & Iteration & Para. & Iteration & Iteration\\
\noalign{\smallskip}\hline\noalign{\smallskip}
$E_1$ & n=22 & 46 & n=20 & 52 & n=20 & 44 & n=12,$\gamma$=1 & 41 & n=15,$\gamma$=3 & 43 & n=12 ,$\gamma$=3  & 43 & n=13 & 51 & 44\\
$E_2$ & n=29 & 61 & n=28 & 66 & n=24 & 54 & n=18,$\gamma$=2 & 52 & n=18,$\gamma$=3 & 51 & n=14 ,$\gamma$=3 & 53 & n=16 & 65 & 51\\
$E_3$ & n=74 & 127 & n=70 & 141 & n=57 & 105 & n=38,$\gamma$=3 & 99 & n=38,$\gamma$=5 & 99 & n=22 ,$\gamma$=5 & 97 & n=41 & 137 & 102\\
$E_4$ & n=18 & 64 & n=32 & 72 & n=13 &52  & n=18,$\gamma$=3 & 50 & n=12,$\gamma$=3 & 54 & n=8 ,$\gamma$=3 & 55 & n=9 & 63 & 54\\
$E_5$ & n=298 & 1086 & n=33 & 880 & n=102 & 646 & n=3,$\gamma$=5 & 351 & n=16,$\gamma$=8 & 426 &  n=1,$\gamma$=7 & 401 & n=9 & 629 & 510\\
$E_6$ & n=653 & 2500 & n=65 & 1677 & n=216 & 1419 & n=6,$\gamma$=9 & 678 & n=32,$\gamma$=12 & 834& n=3,$\gamma$=9 & 772 & n=17 & 1129 & 934\\
$E_7$ & n=2356 & 9613 & n=204 & 5191 & n=820 & 5423 & n=17,$\gamma$=20 & 2032 & n=105,$\gamma$=25 & 2540& n=6,$\gamma$=17 & 2262 & n=59 & 3733 & 2737\\
$E_8$ & n=216 & 783 & n=28 & 754 & n=57 & 432 & n=2,$\gamma$=4 & 298 & n=13,$\gamma$=6 & 325& n=1,$\gamma$=5 & 306 & n=9 & 586 & 538\\
$E_9$ & n=881 & 2363 & n=55 & 1445 & n=326 & 1384 & n=5,$\gamma$=7 & 598 & n=33,$\gamma$=12 & 729& n=3,$\gamma$=8 & 612 & n=24 & 1072 & 735\\
\noalign{\smallskip}\hline
\end{tabular}
\end{sidewaystable}

\begin{table*}
\caption{The parameter tuning cost (number of extra interactions) of the compared algorithms in environments $E_1$ to $E_9$}
\label{tab:cost}       
\begin{tabular}{p{0.02\textwidth}*{7}{p{0.1\textwidth}<{\centering}}}
\hline\noalign{\smallskip}
Env.& DP$_{ri}$ & DGPA & DBPA & DGCPA$^*$ & SE$_{ri}$ & GBSE & LELA$_{R}$ \\
\noalign{\smallskip}\hline\noalign{\smallskip}
$E_1$ & $3.075 \times 10^{6}$ & $3.023 \times 10^{6}$ & $2.523 \times 10^{6}$ & $3.046 \times 10^{7}$ & $2.062 \times 10^{7}$ & $ 1.881 \times 10^{7}$ & $2.471 \times 10^{6}$ \\
$E_2$ & $3.866 \times 10^{6}$ & $4.552 \times 10^{6}$ & $3.373 \times 10^{6}$ & $3.633 \times 10^{7}$ & $2.669 \times 10^{7}$ & $ 2.241 \times 10^{7}$ & $3.346 \times 10^{6}$ \\
$E_3$ & $1.521 \times 10^{7}$ & $1.554 \times 10^{7}$ & $1.045 \times 10^{7}$ & $9.192 \times 10^{7}$ & $8.704 \times 10^{7}$ & $ 6.180 \times 10^{7}$ & $1.042 \times 10^{7}$ \\
$E_4$ & $2.616 \times 10^{6}$ & $5.331 \times 10^{6}$ & $2.147 \times 10^{6}$ & $3.445 \times 10^{7}$ & $2.215 \times 10^{7}$ & $ 2.025 \times 10^{7}$ & $2.362 \times 10^{6}$ \\
$E_5$ & $3.947 \times 10^{8}$ & $6.248 \times 10^{7}$ & $1.033 \times 10^{8}$ & $2.421 \times 10^{8}$ & $1.268 \times 10^{9}$ & $ 3.443 \times 10^{8}$ & $2.437 \times 10^{7}$ \\
$E_6$ & $1.813 \times 10^{9}$ & $1.708 \times 10^{8}$ & $4.117 \times 10^{8}$ & $7.442 \times 10^{8}$ & $6.905 \times 10^{9}$ & $ 9.331 \times 10^{8}$ & $6.262 \times 10^{7}$ \\
$E_7$ & $1.503 \times 10^{10}$ & $1.369 \times 10^{9}$ & $4.931 \times 10^{9}$ & $7.618 \times 10^{9}$ & $1.207 \times 10^{11}$ & $9.158 \times 10^{9}$ & $3.910 \times 10^{8}$ \\
$E_8$ & $2.008 \times 10^{8}$ & $5.264 \times 10^{7}$ & $4.146 \times 10^{7}$ & $1.808 \times 10^{8}$ & $7.079 \times 10^{8}$ & $ 2.714 \times 10^{8}$ & $2.209 \times 10^{7}$ \\
$E_9$ & $1.802 \times 10^{9}$ & $1.208 \times 10^{8}$ & $5.933 \times 10^{8}$ & $5.495 \times 10^{8}$ & $6.266 \times 10^{9}$ & $ 8.092 \times 10^{8}$ & $7.029 \times 10^{7}$ \\
\noalign{\smallskip}\hline
\end{tabular}
\end{table*}

However, taking the parameter tuning cost of the competitors into consideration, the parameter-free property begins to show its superiority. In order to clarify that point, we count the number of interactions between automaton and environment during the process of parameter tuning for each parameter-based scheme. The results are summarized in table \ref{tab:cost}\footnotemark[5]\footnotetext[5]{It is noted that the numerical value shown in Table \ref{tab:cost} may differ according to the way parameter tuning being implemented, still it gives qualitatively evidence to the heavy parameter tuning cost of the parameter-based schemes. The technical details of parameter tuning procedure used here is provided in the Appendix.}. It can be seen that the extra interactions required for parameter tuning by deterministic estimator based schemes (DGPA, DBPA and LELA$_R$, except DP$_{ri}$) are slightly less than stochastic estimator based schemes (DGCPA$^{*}$, SE$_{ri}$ and GBSE). Both families of schemes cost millions of extra interactions for seeking for the \emph{best parameter}. The proposed scheme can achieve a comparative performance without relying on any extra informations.

\subsection{Comparison with Untuned Schemes} 
	In this part, the parameter-based algorithms are simulated in benchmark environments without their learning parameter specifically tuned. Their performance will be compare with PFLA under the same condition -- no extra information about the environment are available.

\subsubsection{Using Generalized Learning Parameter}
Firstly, the best parameter in $E_2$ and $E_6$ are applied for learning in other environments respectively to evaluate how well they can `generalize' in other environments. The results are shown in Table \ref{tab:generalization2} and Table \ref{tab:generalization6} respectively.

\begin{table*}
\caption{Comparison of convergence rate and accuracy of the parameter-based algorithms in all environments other tha $E_2$, when using the `best' learning parameters in $E_2$}
\label{tab:generalization2}       
\begin{tabular}{{p{0.01\textwidth}<{\centering}}*{14}{p{0.045\textwidth}<{\centering}}}
\hline\noalign{\smallskip}
\multirow{2}{*}{\centering Env.} & \multicolumn{2}{c}{DP$_{ri}$} & \multicolumn{2}{c}{DGPA} & \multicolumn{2}{c}{DBPA} & \multicolumn{2}{c}{DGCPA*} & \multicolumn{2}{c}{SE$_{ri}$} & \multicolumn{2}{c}{GBSE}& \multicolumn{2}{c}{LELA$_{R}$}  \\
& Iteration & Accuracy & Iteration & Accuracy& Iteration & Accuracy & Iteration & Accuracy& Iteration & Accuracy& Iteration & Accuracy& Iteration & Accuracy\\
\noalign{\smallskip}\hline\noalign{\smallskip}
$E_1$ & 55 & 0.998 & 65 & 0.999 & 49 & 0.998 & 53 & 0.995 & 47 & 0.999 & 48 & 0.999 & 59 & 0.998 \\
$E_3$ & 63 & 0.975 & 70 & 0.976 & 57 & 0.976 & 61 & 0.972 & 57 & 0.979 & 62 & 0.983 & 67 & 0.976 \\
$E_4$ & 90 & 0.999 & 66 & 0.996 & 79 & 0.999 & 45 & 0.994 & 69 & 0.999 & 80 & 0.999 & 96 & 0.999 \\
$E_5$ & 264 & 0.895 & 767 & 0.995 & 301 & 0.967 & 640 & 0.999 & 286 & 0.962 & 701 & 0.998 & 1026 & 0.999 \\
$E_6$ & 319 & 0.804 & 835 & 0.971 & 408 & 0.927 & 821 & 0.996 & 351 & 0.897 & 836 & 0.987 & 1068 & 0.995 \\
$E_7$ & 393 & 0.658 & 976 & 0.858 & 577 & 0.806 & 1249 & 0.957 & 443 & 0.748 & 1093 & 0.905 & 1155 & 0.909 \\
$E_8$ & 253 & 0.918 & 752 & 0.997 & 280 & 0.979 & 616 & 0.999 & 273 & 0.981 & 648 & 0.999 & 954 & 0.999 \\
$E_9$ & 246 & 0.801 & 827 & 0.981 & 275 & 0.869 & 751 & 0.997 & 299 & 0.903 & 636 & 0.984 & 758 & 0.986 \\
\noalign{\smallskip}\hline
\end{tabular}
\end{table*}

\begin{table*}
\caption{Comparison of convergence rate and accuracy of the parameter-based algorithms in all environments other tha $E_6$, when using the `best' learning parameters in $E_6$}
\label{tab:generalization6}       
\begin{tabular}{{p{0.01\textwidth}<{\centering}}*{14}{p{0.045\textwidth}<{\centering}}}
\hline\noalign{\smallskip}
\multirow{2}{*}{\centering Env.} & \multicolumn{2}{c}{DP$_{ri}$} & \multicolumn{2}{c}{DGPA} & \multicolumn{2}{c}{DBPA} & \multicolumn{2}{c}{DGCPA*} & \multicolumn{2}{c}{SE$_{ri}$} & \multicolumn{2}{c}{GBSE}& \multicolumn{2}{c}{LELA$_{R}$}  \\
& Iteration & Accuracy & Iteration & Accuracy& Iteration & Accuracy& Iteration & Accuracy& Iteration & Accuracy& Iteration & Accuracy& Iteration & Accuracy\\
\noalign{\smallskip}\hline\noalign{\smallskip}
$E_1$ & 812 & 1 & 125 & 0.999 & 282 & 1 & 29 & 0.783 & 95 & 1 & 26 & 0.769 & 61 & 0.999 \\
$E_2$ & 923 & 1 & 126 & 0.999 & 320 & 1 & 28 & 0.777 & 103 & 0.999 & 29 & 0.781 & 68 & 0.998 \\
$E_3$ & 899 & 1 & 133 & 0.996 & 317 & 1 & 31 & 0.725 & 124 & 0.998 & 29 & 0.716 & 70 & 0.978 \\
$E_4$ & 1572 & 1 & 126 & 0.999 & 535 & 1 & 28 & 0.791 & 137 & 1 & 46 & 0.846 & 101 & 0.999 \\
$E_5$ & 1879 & 0.999 & 1582 & 0.999 & 969 & 0.999 & 501 & 0.972 & 641 & 0.999 & 599 & 0.999 & 1085 & 0.999 \\
$E_7$ & 3961 & 0.942 & 1939 & 0.945 & 2358 & 0.965 & 1055 & 0.929 & 1203 & 0.942 & 1110 & 0.948 & 1219 & 0.917 \\
$E_8$ & 1641 & 0.999 & 1555 & 0.999 & 845 & 0.999 & 495 & 0.974 & 629 & 0.999 & 592 & 0.999 & 1008 & 0.999 \\
$E_9$ & 1923 & 0.987 & 1667 & 0.998 & 1078 & 0.988 & 673 & 0.973 & 725 & 0.996 & 675 & 0.997 & 799 & 0.989 \\
\noalign{\smallskip}\hline
\end{tabular}
\end{table*}

\subsubsection{Using Random Learning Parameter}
Secondly, randomly selected learning parameters are adopted to evaluate the expected performance of each algorithm in fully unknown environments. The random resolution parameter takes value in the range from is 90\% of the minimal value to 110\% of the maximal value of the best resolution parameter in the nine benchmark environment\footnotemark[5]\footnotetext[5]{For example, the resolution parameter of DPri is range from $\lfloor90\%*18\rfloor$ to $\lceil110\%*2356\rceil$, i.e, from 16 to 2592}, and a range from 1 to 20 for the perturbation parameter if needed. The simulation results are demonstrated as Table \ref{tab:random}.

\begin{table*}
\caption{Comparison of the average number of iterations required for convergence of the parameter-based algorithms in environments $E_1$ to $E_9$. The randomly selected learning parameters are used and 250,000 experiments were performed for each scheme in each environment}
\label{tab:random}       
\begin{tabular}{{p{0.01\textwidth}<{\centering}}*{14}{p{0.045\textwidth}<{\centering}}}
\hline\noalign{\smallskip}
\multirow{2}{*}{\centering Env.} & \multicolumn{2}{c}{DP$_{ri}$} & \multicolumn{2}{c}{DGPA} & \multicolumn{2}{c}{DBPA} & \multicolumn{2}{c}{DGCPA*} & \multicolumn{2}{c}{SE$_{ri}$} & \multicolumn{2}{c}{GBSE}& \multicolumn{2}{c}{LELA$_{R}$}  \\
& Iteration & Accuracy & Iteration & Accuracy& Iteration & Accuracy& Iteration & Accuracy& Iteration & Accuracy& Iteration & Accuracy& Iteration & Accuracy\\
\noalign{\smallskip}\hline\noalign{\smallskip}
$E_1$ & 1606 & 0.999 & 216 & 0.999 & 574 & 0.999 & 76 & 0.924 & 121 & 0.999 & 71 & 0.943 & 108 & 0.999 \\
$E_2$ & 1824 & 0.999 & 217 & 0.999 & 652 & 0.999 & 73 & 0.922 & 132 & 0.999 & 78 & 0.943 & 121 & 0.999 \\
$E_3$ & 1767 & 0.999 & 224 & 0.996 & 638 & 0.998 & 87 & 0.896 & 152 & 0.996 & 95 & 0.927 & 124 & 0.990 \\
$E_4$ & 3121 & 0.999 & 217 & 0.999 & 1105 & 0.999 & 66 & 0.928 & 189 & 0.999 & 109 & 0.953 & 192 & 0.999 \\
$E_5$ & 3253 & 0.996 & 2821 & 0.999 & 1476 & 0.997 & 836 & 0.977 & 687 & 0.993 & 925 & 0.997 & 2153 & 0.999 \\
$E_6$ & 3995 & 0.988 & 2922 & 0.995 & 2072 & 0.993 & 1042 & 0.975 & 886 & 0.979 & 1158 & 0.991 & 2229 & 0.996 \\
$E_7$ & 6260 & 0.951 & 3309 & 0.963 & 3626 & 0.970 & 1647 & 0.952 & 1302 & 0.911 & 1699 & 0.952 & 2395 & 0.958 \\
$E_8$ & 2859 & 0.997 & 2774 & 0.999 & 1288 & 0.999 & 810 & 0.978 & 647 & 0.996 & 878 & 0.998 & 2003 & 0.999 \\
$E_9$ & 3031 & 0.985 & 2919 & 0.997 & 1615 & 0.987 & 1041 & 0.976 & 768 & 0.979 & 988 & 0.988 & 1547 & 0.993 \\
\noalign{\smallskip}\hline
\end{tabular}
\end{table*}

From the three tables, there is a significant decline of accuracy in some environments. As the accuracies differ greatly in those cases, the convergence rates cannot be compared directly. 
However, several conclusions can be drawn. One is that the performance of untuned parameter-based algorithms is unstable when learning in unknown environment, thus cannot be used in practical applications without parameter tuning. Another conclusion is that those algorithms, who use generalized learning parameters or random learning parameters, are either have a lower accuracy or a slower convergence rate than PFLA in the benchmark environment. In other words, none of them can outperform PFLA in both accuracy and convergence rate without the help of prior information.

\subsection{Discussion of the Fairness of the Comparison}
Technically speaking, the comparison between PFLA and well-tuned schemes are not fair. This is because that the interactions can be perceived as information exchanges between automaton and the environment. So if the number of interaction is unlimited, the algorithm can simply use the empirical distributions. The outperforming of the well-tuned schemes owes to their richer knowledge about the environment acquired during the parameter tuning process. And for this reason, a fair comparison between PFLA and untuned schemes is carried out. Despite the unfairness of the first comparison, the significance lies in providing baselines for evaluating the convergence rate of PFLA qualitatively .

By the way, the comparison within parameter-based algorithms is not fair either, because the amount of prior information acquired is different. This method is be widely used by the research community to compare the theoretically best performance of their proposed algorithms, however the hardness of the algorithm can achieve theoretically best is usually ignored.

\section{Conclusion}
\label{sec:conclusion}
In this paper, we propose a parameter-free learning automaton scheme for learning in stationary stochastic environments. The proof of the $\varepsilon$-optimality of the proposed scheme in every stationary random environments  is presented. Compared with existing schemes, the proposed PFLA possesses a parameter-free property, i.e, a set of parameters can be universally applicable for all environments. Furthermore, our scheme is evaluated in four two-action and five ten-action benchmark environments and compared with several classic and the state-of-art schemes in the field of LA. Simulations confirm that our scheme can converge to the optimal action with a high accuracy. Although the rate of convergence is outperformed by some schemes that are well tuned for specific environments, the proposed scheme still shows its intriguing property of not relying on parameter-tuning process. Our future work includes optimizing the exploration strategy further.  

\appendix[The Standard Parameter Tuning Procedure of Learning Automata]
\label{tuning}
As emphasized in section \ref{sec:introduction}, parameter tuning is intend to balance the trade-off between speed and accuracy. And the standard procedure of parameter tuning is pioneered in \cite{oommen1990discretized} and become a common practive in follow-up researches\cite{oommen2001continuous,agache2002generalized,papadimitriou2004new,zhang2014last,jiang2015new,ge2015novel}.

The basic idea is, the smallest value of resolution parameter $n$ that yielded the fastest convergence and that simultaneously resulted in zeros errors in a sequence of $NE$ experiments are defined as ``best'' parameters. Besides, to reduce the variance coefficient of the ``best'' values of $n$, \cite{papadimitriou2004new} advocate to perform the same procedure 20 times, and compute the average ``best'' value of $n$ in these experiments. For tuning stochastic estimator based learning automata, who has a perturbation parameter $\gamma$ in addition to the well known resolution parameter $n$. A two-dimensional grid search should be performed to seek for the best parameter pair $(n,\gamma)$. The method used in \cite{papadimitriou2004new} is to obtain ``best'' resolution parameter $n$ for each value of $\gamma$, and then evaluate the speed of convergence for each of the $(n,\gamma)$ pairs and choose the best pair.

Based on these instructions, we use the following procedure for parameter tuning in our experiment:

The resolution parameter is initialized to 1, and increased by 1 each time a wrong converge emerging until a certain number of successive \emph{No Error} experiments is carried out. Repeat this process 20 times, averaging over these 20 resulting values and denote it as the  best resolution parameter. The value of number of successive No Error experiments is set as $NE=750$, as the same value in \cite{papadimitriou2004new,zhang2014last,jiang2015new,ge2015novel}. For tuning the ``best'' $\gamma$, In our simulation settings, for the four two-action environments, the search range of $\gamma$ is from 1 to 10; For the five ten-action environments except $E_7$, the search range of $\gamma$ is from 1 to 20, while for $E_7$, the most difficult one, the range is a little wider, from 1 to 30.

It is noted that the above standard procedure have been widely adopted by the research community, but it does not mean this is the most efficient way. Apparently, it can be improved by several methods, such as random search or two-stage coarse-to-fine search. This issue is worth further investigation and is beyond the scope of this paper.


%





\ifCLASSOPTIONcaptionsoff
  \newpage
\fi



\bibliographystyle{IEEEtran}
\bibliography{bare_jrnl}
\end{document}